\documentclass[11pt]{article}
\usepackage{coling2018}
\usepackage{times}
\usepackage{url}
\usepackage{latexsym}
\usepackage{graphicx}
\usepackage[export]{adjustbox}

\usepackage{enumitem}
\usepackage{tabularx}
\usepackage{nicefrac}
\usepackage{multirow}
\usepackage{hyperref}
\usepackage{xstring}
\usepackage{color}
\RequirePackage{colortbl}

\newcolumntype{P}[1]{>{\centeringarraybackslash}p{#1}}

\title{WikiRef: Wikilinks as a route to recommending\\ appropriate references for scientific Wikipedia pages}
\author{Abhik Jana \\
 IIT Kharagpur \\
  Kharagpur, India \\
  {\tt abhik.jana@iitkgp.ac.in} \\\\
   \textbf{Pawan Goyal} \\
IIT Kharagpur \\
  Kharagpur, India \\
  {\tt pawang@cse.iitkgp.ac.in} \\\And
 Pranjal Kanojiya \\
 IIT Kharagpur \\
  Kharagpur, India \\
  {\tt pranjal989091@gmail.com } \\\\
  \textbf{Animesh Mukherjee} \\
IIT Kharagpur \\
  Kharagpur, India \\
  {\tt animeshm@gmail.com} \\}
  
\date{}

\begin{document}
\maketitle
\begin{abstract}
The exponential increase in the usage of Wikipedia as a key source of scientific knowledge among the researchers is making it absolutely necessary to metamorphose this knowledge repository into an integral and self-contained source of information for direct utilization. Unfortunately, the references which support the content of each Wikipedia entity page, are far from complete. Why are the reference section ill-formed for most Wikipedia pages? Is this section edited as frequently as the other sections of a page? Can there be appropriate surrogates that can automatically enhance the reference section? In this paper, we propose a novel two step approach -- \textbf{WikiRef} -- that (i) leverages the wikilinks present in a scientific Wikipedia target page and, thereby, (ii) recommends highly relevant references to be included in that target page appropriately and automatically borrowed from the reference section of the wikilinks. In the first step, we build a classifier to ascertain whether a wikilink is a potential source of reference or not. In the following step, we recommend references to the target page from the reference section of the wikilinks that are classified as potential sources of references in the first step. We perform an extensive evaluation of our approach on datasets from two different domains -- Computer Science and Physics. For Computer Science we achieve a notably good performance with a \textit{precision@1} of 0.44 for reference recommendation as opposed to 0.38 obtained from the most competitive baseline. For the Physics dataset, we obtain a similar performance boost of 10\% with respect to the most competitive baseline. 
\end{abstract}

\section{Introduction}
\label{intro}

%
%
\blfootnote{
    %
    %
    %
    
     \hspace{-0.65cm}  
     This work is licenced under a Creative Commons 
     Attribution 4.0 International Licence.
     Licence details:
     \url{http://creativecommons.org/licenses/by/4.0/}
     
    %
}
Wikipedia, the largest online encyclopedia, is a collaborative work of 33,778,487 users and 1,210 admins, with 5,662,889 English articles and a total of 45,132,517 articles in more than 250 languages as of June 7, 2018\footnote{https://en.wikipedia.org/wiki/Wikipedia:Size\_of\_Wikipedia}. Introduced in 2001, it has become one of the most visited websites having global Alexa Rank -- 5\footnote{http://www.alexa.com/siteinfo/www.wikipedia.org}. Each Wikipedia article contains a rich collection of concepts along with a set of hyperlinks associating crucial terms to other wiki pages (termed as wikilinks), which makes the article more comprehensive. In addition, the ability to edit the entity pages easily with proper source citations and to view the edits reflected in the pages timely are some of the key reasons behind Wikipedia's growth and popularity. As a consequence, Wikipedia has become a destination of all kinds of information about entities and events. Further, Wikipedia is being widely utilized by many applications focused on entity linking and disambiguation~\cite{moro2014entity,hachey2013evaluating,cucerzan2007large,hoffart2011robust}, named entity disambiguation~\cite{cucerzan2007large,barrena2015combining}, semantic similarity~\cite{gabrilovich2007computing,wu2015sense}, information extraction~\cite{wu2010open} etc. 

\noindent{\bf What is lacking?} Nevertheless, there are different competing opinions among the researchers regarding the reliability, integrity and usage of Wikipedia as an authentic source of scientific information. While many researchers consider Wikipedia as a valuable resource, others question the accuracy, comprehensiveness and completeness of the articles. For instance, according to ~\newcite{gorman2007tale}, Wikipedia is an ``unethical resource unworthy of our respect'' whereas~\newcite{orlowski2005wikipedia} questions that the technology which makes Wikipedia possible is not a substitute for expert editors, and poor writing will remain unavoidable without any expertise.~\newcite{chesney2006empirical} conducted an experiment to evaluate the credibility of Wikipedia and found that the studies focusing only on history articles provide mixed evidence concerning its accuracy. The general solution is to post-process the edits done by the users. In order to ensure the accuracy and quality of writing of articles, Wikipedia has imposed a new policy of requiring senior editors to approve changes to articles. 

Even then, the completeness of the article is not fully ensured. Not all Wikipedia pages referring to entities (entity pages) are comprehensive: relevant information can either be missing or added with a delay. In addition, the frequency of editing of different sections in a given Wikipedia article varies extensively. For instance, frequency of editing text is higher than frequency of editing wikilinks or references; also add/edit references is considered as the hardest task\footnote{\url{https://www.mediawiki.org/wiki/Editing_Tasks_Survey}}. This opens up the requirement for an automated system which increases the number of relevant references by either processing the content of the target article itself or by appropriately collating references from alternative sources so as to make the overall information content of the article more complete.

\noindent{\bf State-of-the-art}: Efforts have been made by the researchers to populate Wikipedia pages automatically.~\newcite{sauper2009automatically} propose an approach for populating Wikipedia pages with content coming from external sources by automatically generating whole entity pages for specific entity classes.~\newcite{taneva2013gem} propose an approach to generate novel summaries from the external information which could be added to Wikipedia entity pages.~\newcite{balog2013cumulative} and ~\newcite{balog2013multi} try to recommend news citations for an entity in Wikipedia.~\newcite{west2014knowledge} focuses on the problem of knowledge base completion, through question answering and tries to complete missing facts in Freebase based on templates. All these works significantly rely on high quality input sources which are utilized to extract textual facts for Wikipedia page population. In one of the recent works,~\newcite{fetahu2015automated} propose a two-stage supervised approach for suggesting news articles corresponding to entity pages. Attempt has also been made to introduce appropriate wikilinks automatically~\cite{ikikat2015automatic}. In a similar line,~\newcite{raganato2016automatic} present the automatic construction and evaluation of a Semantically Enriched Wikipedia (SEW) in which the overall number of wikilinks has been more than tripled. Efforts have also been made to generate Wikipedia articles of named entities in a semi-supervised framework~\cite{Pochampally2016SemiSupervisedAG}.

\noindent\textbf{Motivation}: Recently, with the increasing number of scientific articles and with the need for getting an overall view of a particular scientific topic within a very less amount of time, researcher's tendency of referring to the Wikipedia pages instead of going through a set of very specific scientific articles is massively increasing. Given that, such a huge community of researchers rely on Wikipedia for scientific information, it becomes absolutely necessary to deeply focus on the completeness of the Wikipedia articles which talk about some scientific topics. 

\begin{figure}[!tbh]
\vspace{-2mm}
    \centering
    \includegraphics[width=0.8 \textwidth]{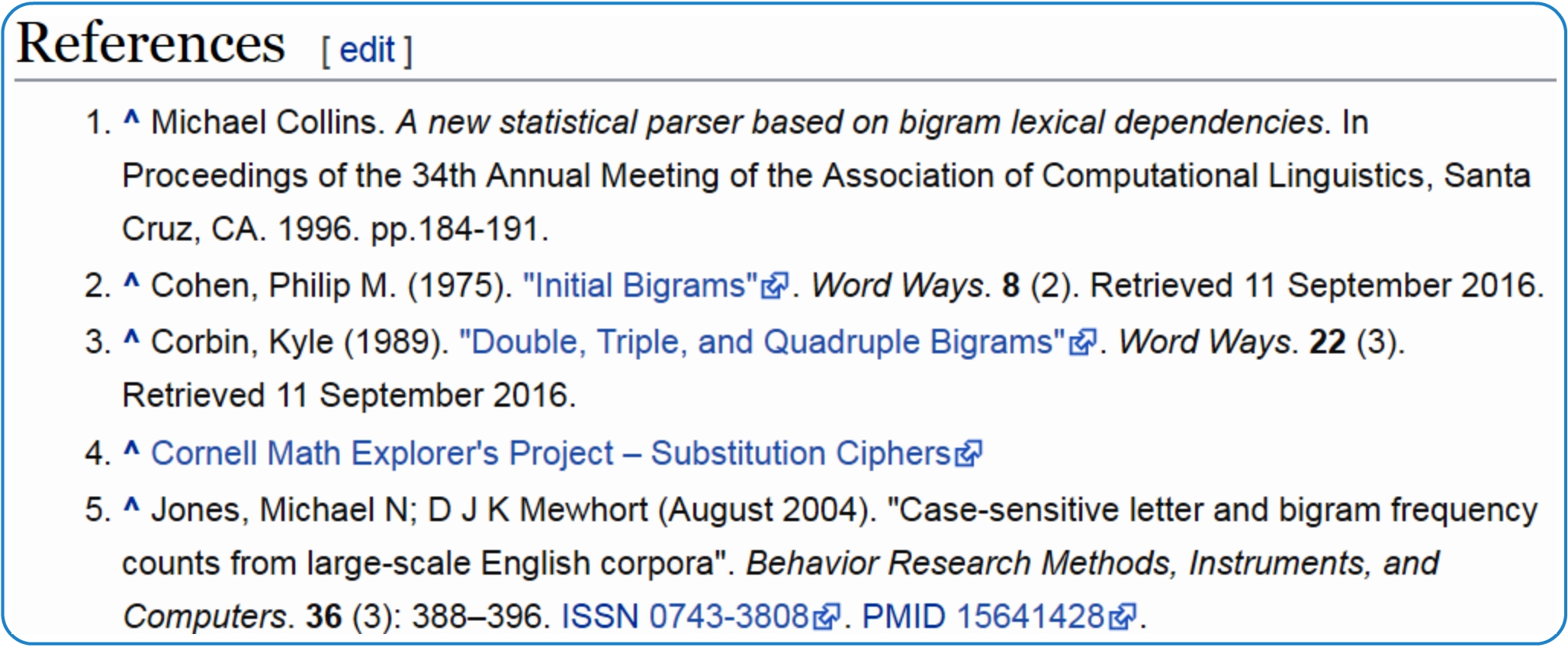}
 \vspace{-4mm}
    \caption{Snapshot of Wikipedia `Bigram' in the year 2018.
    }
    \label{fig:Bigram}
    \vspace{-2mm}
\end{figure}

In the following we enlist some initial observations to show that there is a huge scope of improvement of the scientific Wikipedia articles through the inclusion of a proper set of references. Figure~\ref{fig:Bigram} illustrates the snapshot of the reference section of the `Bigram' page\footnote{\url{https://en.wikipedia.org/wiki/Bigram}} in the year 2018. There are only five references present in the page; very relevant scientific articles like, ``Statistical Identification of Language''~\cite{dunning1994statistical}, ``Foundations of Statistical Natural Language Processing''~\cite{manning1999foundations}, etc. are missing, without which the article seems to be incomplete. Also from the survey conducted by wikimedia~\footnote{\url{https://commons.wikimedia.org/wiki/File:WMF_editing_tasks_survey_July_2015.pdf}} 
it has been observed that adding or editing the reference section is harder compared to adding or editing simple text or a wikilink which therefore leads to much less frequent changes of the reference section. In order to further confirm this trend, we conduct an analysis on the edit history of 1120 Wikipedia articles from the Computer Science category and observe that till Jan, 2017 on average around 65\% edits are in the text content, 32\%  are wikilink edits and strikingly only 1\% are reference edits. Rest correspond to table, template and category edits etc. We present some example Wikipedia articles along with the statistic of their edit history in Table~\ref{tab:examples_eh}. Clearly, all these observations point to the requirement of an automated reference recommendation system to keep the reference section of a scientific Wikipedia article up-to-date.

\begin{table}[!tbh]
\vspace{-2mm}
\begin{center}
    \begin{tabular}{ |p{3.3cm}|p{1.5cm}|p{1.5cm}|p{2.2cm}|p{2.8cm}|p{2.1cm}|}
    \hline
   \centering Wikipage title & \centering Start date & \centering End date & \centering \#(Total edits) & \centering \#(Wikilink edits) & \centering \#(Ref. edits)\tabularnewline
   \hline
    \centering Connection Machine & \centering 7/8/02 & \centering 6/1/17 & \centering 242 & \centering 98 & \centering 3
  \tabularnewline \hline
	    \centering Fuzzy Logic & \centering 15/4/02 & \centering 29/1/17 & \centering 1343 & \centering 390 & \centering 5  \tabularnewline \hline
        \centering Javascript & \centering 5/2/02 & \centering 31/1/17 & \centering 4141 & \centering 1146 & \centering 69        
  \tabularnewline \hline

\end{tabular}
\end{center}
\vspace{-4mm}
\caption{Representative scientific Wikipedia articles with detailed statistic of their edit history. Start date indicates the date of creation of Wikipage and end date indicates the time point till which we perform the analysis of edit history.}
\label{tab:examples_eh}
\vspace{-2mm}
\end{table}

\noindent{\bf What we envisage?} Motivated by the fact that wikilink edits are less hard and more frequent compared to reference edits, a natural question that arises is that whether it is possible to develop an automatic procedure to add a set of references whenever there is a wikilink edit on a target page. In this paper, we investigate the possibility of enriching the reference section of a scientific Wikipedia article using the wikilinks present in the article as a surrogate. In particular, \textit{we introduce the novel idea of reference inheritance from the source pages pointed to by the wikilinks present on the target page to populate the reference section of the target page with relevant references}.

For this study, we consider 3842 target Wikipedia pages (till June, 2017) from the Computer Science domain. In order to show that our method is quite generic, we also consider an additional dataset comprising 2871 target Wikipedia pages (till February, 2018) from the Physics domain.

We propose a novel two step approach as follows. In the first step, \textit{we determine the potential wikilinks from which references could be inherited}. In the second step, \textit{we recommend the most relevant references from the potential wikilinks obtained in the first step}.

\noindent{\bf Key contributions}: In summary, we make the following contributions: \\
a) \textit{Novel problem formulation}: We formulate and address the problem of automatically populating the reference section by inheriting the references from the wikilinks present in a target scientific Wikipedia article.\\   
b) \textit{The two step approach}: We propose a two-step reference recommendation system and show its advantage over simple baselines that do not implement the first step of potential wikilink selection from which references could be inherited.\\
c) \textit{Evaluation}: We perform extensive experiments to evaluate our system. We evaluate in two different ways - first, by comparing our recommendations with the references already present on the target page (acting as a ground-truth) and second, through human judgement where experts are tasked to judge the appropriateness of the recommended references. For the automatic evaluation we obtain a \textit{precision@1} of 0.44 for the Computer Science dataset compared to 0.38 obtained for the most competitive baseline. For the Physics dataset we obtain a similar performance boost of 10\% over the best baseline. Human judgement experiments show that the average Spearman's rank correlation coefficient ($\rho$) between the ranked list returned by WikiRef and the one produced by averaging the responses of the annotators is \textbf{0.203}  compared to 0.168 obtained for the most competitive baseline\footnote{We perform the human judgement experiment only for the CS domain because of the background of the contributing authors of this paper.}.

\section{Problem formulation}
\label{PF}
Let us assume that we have a scientific Wikipedia article $A$ (i.e., target) whose reference section we wish to populate using a set of relevant references. As an input we use only the text content and wikilinks present in $A$. Suppose, $A$ has a set of $n$ wikilinks $B_1, B_2, \dots, B_n$ which in this case simply correspond to the $n$ Wikipedia source articles from which $A$ could possibly inherit the references. We assume that while the reference sections of the individual source pages themselves may not be ``well populated'', together the reference sections of all the source articles may provide a potential set of references, relevant to $A$. We divide this task in two major steps. First, we obtain the wikilinks (i.e., a set of $B$s) that are appropriate for inheriting references from. We pose this as a binary classification problem thus dividing the set of $B$s in those that are relevant (appropriate for inheriting references) vs those that are not. Second, we prepare a ranked list of $k$ references taken from each of the correctly classified (i.e., relevant) wikilinks which could be added to the reference section of $A$. We pose this second step as a ``learning-to-rank'' problem and use various features to prepare the final list of recommendations.

\section{Dataset}
\label{ds}
We use datasets from two different domains -- Computer Science (CS) and Physics (PH) -- to evaluate our system. Note that while CS is our primary focus (owing to the background of the authors) we use the PH dataset to show that our method is quite generic.

\noindent\textbf{CS dataset}: In order to prepare the CS dataset we crawl a set of 3842 target Wikipedia pages till June, 2017. The topics we span in this crawl are information retrieval, machine learning, automata theory, graph theory  etc. In addition, we also crawl all the source pages corresponding to all the wikilinks present in the 3,842 target articles. The total number of wikipages thus crawled is 121,154.

\noindent\textbf{PH dataset}: For the PH dataset we crawl a total of 2,871 target articles spanning topics like gravitation, mechanics, motion etc. The source pages pointed to by the wikilinks present on these target articles together make the total crawl size of 107,332 pages.


\section{Building WikiRef}
Recall, that our approach works in two steps. We present the details of each of these steps below which together refers to our system -- WikiRef.

\subsection{Classification of wikilinks (Step - I)}
\label{sec:CW}

As discussed in Section~\ref{PF}, we have a target Wikipedia article $A$ with $n$ wiklinks ($B_1, B_2, \dots, B_n$) present in it. Our goal in this step is to ascertain how appropriate a source page pointed to by a wikilink $B_i$ is for inheriting references. To estimate this, we measure the extent of similarity between $A$ and the page pointed to by $B_i$. The hypothesis is that the higher this similarity, the  more relevant is the source page to $A$ and thus higher is the propensity of reference inheritance. We define various notions of similarity that act as features to a binary classifier which classifies if the source is relevant or not.

\begin{itemize}[noitemsep,nolistsep,leftmargin=*]
\item {\bf Tf-idf similarity of article summaries (TIS):} We represent the summaries (first paragraphs) of $A$ and the source page pointed to by $B_i$ as tf-idf vectors and compute the cosine similarity between them. Note that, inverse document frequency (idf) is calculated only from the dataset we have prepared.
\item{\bf Outlinks similarity (OS)}: Outlinks of a Wikipedia article (say $A$) is the set of Wikipedia articles that $A$ hyperlinks to. We compute outlinks similarity as the Jaccard overlap between the outlinks of $A$ and the source page pointed to by $B_i$.
\item{\bf Inlinks similarity (IS)}: Inlinks of a Wikipedia article (say $A$) is the set of Wikipedia articles that hyperlink to the page $A$. We compute inlinks similarity as the Jaccard overlap between the inlinks of $A$ and the source page pointed to by $B_i$. Note that, for computing inlinks we consider the articles present only in our dataset.
\item{\bf Out sentence similarity (OSS)}: Here we consider the common outlinks of $A$ and the source page pointed to by $B_i$. We collate all the sentences in $A$ where these common outlinks occur, to prepare a document $OS_A$. Similarly, we collate all sentences in the page pointed to by $B_i$ where these common outlinks occur to form a document $OS_{B_i}$. Now we represent $OS_A$ and $OS_{B_i}$ as tf-idf vectors and compute cosine similarity between them.
\item{\bf In sentence similarity (ISS)}: This is same as OSS with the exception that here we construct the documents based on the sentences where the common inlinks of $A$ and the page pointed to by $B_i$ occur.
\end{itemize}

\noindent\textbf{Additional deep neural representations}: Apart from these five similarity measures we use additional variants of TIS, OSS and ISS where instead of representing a text document as a tf-idf vector we represent it as a document vector obtained using an encoder based on a bi-directional LSTM architecture with max pooling, trained on the Stanford Natural Language Inference (SNLI) dataset~\cite{bowman2015large} 
as proposed by~\newcite{conneau-EtAl:2017:EMNLP2017}. Note that, for applying this architecture we use the GloVe vectors trained on Common Crawl data (840B tokens)\footnote{\url{https://nlp.stanford.edu/projects/glove/}}
as seeds for representing words in a document. We name these variants of TIS, OSS and ISS respectively as vector similarity of article summaries (VS), out-sentence vector similarity (OSVS) and in-sentence vector similarity (ISVS) respectively.

We plug in all these features in classifiers like SVM, Random Forest, Logistic Regression etc. We observe that Random Forest performs the best among these and therefore report the results for this classifier only. This step automatically identifies the wikilinks that are potential candidates for reference inheritance.     

\subsection{Recommending the exact list of references (Step - II)}
\label{Sec-s2}
As an outcome of the previous step, we have a target Wikipedia article $A$ and a wikilink $B_i$ which is classified as either relevant for reference inheritance or not. Let us assume that for a $B_i$ that is classified as relevant in the first step, the source page pointed to by $B_i$ has $m$ references $R_1, R_2,...,R_m$. However, all of these $m$ references might not be appropriate to inherit.  Therefore, the task here is to produce a ranked list of $k$ references which $A$ can inherit depending on the relevance of the reference with respect to the content of $A$. In order to estimate the relevance of a reference (say $R_j$) we propose the following features. 

\begin{itemize}[noitemsep,nolistsep,leftmargin=*]
\item{\bf Similarity between the citation context of the reference $R_j$ in the source page pointed to by $B_i$ and the citation context of $B_i$ in $A$}: Citation context represents the sentence in which a reference or wikilink gets cited. We represent the citation contexts as simple tf-idf vectors and compute the cosine similarity between the tf-idf vectors of the citation context of $R_j$ in the source page pointed to by $B_i$ and the citation context of $B_i$ in $A$. We term this feature as \textbf{F1-TI}. We also represent citation contexts as sentence vectors using the bi-LSTM architecture proposed by~\newcite{conneau-EtAl:2017:EMNLP2017} (already discussed earlier) and compute the cosine similarity of these vector representations of the citation contexts of $R_j$ in the source page pointed to by $B_i$ and the citation context of $B_i$ in $A$. We call this feature \textbf{F1-Vec}.     
\item{\bf Similarity between the title of the reference $R_j$  in the source page pointed to by $B_i$ and the citation context of $B_i$ in $A$}:
As the title of any reference contains the most important clue about the reference itself, we represent the title of the reference $R_j$ in the source page pointed to by $B_i$ and citation context of $B_i$ in $A$ as tf-idf vectors and compute the cosine similarity between them. We call this feature \textbf{F2-TI}\footnote{We do not use the corresponding deep vectors since the reference title are in many cases very small and do not produce appropriate vectors thus negatively affecting the overall performance of our system.}.
\end{itemize}

We use -- SVMRank~\cite{joachims2006training} -- a standard learning-to-rank framework and plug in the above three features to obtain a ranked list of $k$ references to recommend. We report the performance our approach for different values of $k$ in Section~\ref{sec:eval}.

\section{Evaluation}
\label{sec:eval}
We first individually evaluate Step - I and Step - II and subsequently report the performance of WikiRef as a whole. We evaluate the performance using the standard precision-recall metrics and compare it with various baseline approaches that we ourselves propose since there is no known work in the literature that could serve as a suitable baseline for this task. 

\subsection{Gold standard dataset}
As discussed in section~\ref{ds} we have a total of 3,842 target Wikipedia pages from the Computer Science (CS) domain and 2,871 target Wikipedia pages from the Physics (PH) domain for which we aim to come up with a list of references. We consider the existing references in the target pages (i.e., the $A$ pages) at the time-point of crawling as the gold standard references to be evaluated against. 

\subsection{Baselines}
There have been several reference recommendation systems~\cite{huang2015neural,caragea2013can,huang2014refseer,tang2014cross,ren2014cluscite,meng2013unified,huang2012recommending} 
for scientific articles but all of them either use the the text of the cited paper or the underlying citation network which makes these systems inappropriate to be used as baselines to compare with WikiRef. Therefore, we propose some standard baselines to ascertain the importance of each step in our approach. 

\noindent\textbf{Baseline I}:  Here, Step - I of our approach is skipped and we assume that all the wikilinks present in the target Wikipedia article are relevant for reference inheritance. In Step - II we rank the references of each wikilink only on the basis of tf-idf similarity between references' citation context in the page pointed to by the wikilink and the wikilink's citation context in target Wikipedia article.

\noindent\textbf{Baseline II}: Here again, Step - I of our approach is skipped and we assume that all the wikilinks present in the target Wikipedia article are relevant for reference inheritance. In Step - II we rank the references of each wikilink only on the basis of tf-idf similarity between the references' title and the wikilink's citation context in the target Wikipedia article.

\noindent\textbf{Baseline III}: Step - I is fully retained. In Step - II we rank the references of each wikilink only on the basis of tf-idf similarity between references' citation context in wikilink's content and wikilink's citation context in target Wikipedia article (F1-TI).

\noindent\textbf{Baseline IV}: Step - I is fully retained. In Step - II we rank the references of each wikilink only on the basis of tf-idf similarity between references' title and wikilink's citation context in the target Wikipedia article (F2-TI).

\noindent\textbf{Baseline V}: Step - I is fully retained. In Step - II we rank the references of each wikilink only on the basis of cosine similarity between sentence vector representation of references' citation context in the page pointed to by the wikilink and the wikilink's citation context in the target Wikipedia article (F1-Vec).


We compare the performance of these baselines with WikiRef in section~\ref{bs}.

\subsection{Performance analysis of Step - I}

We use 70\% of the Wikipedia pages for training and rest 30\% of them for testing using the gold standard dataset. However,  we observe from the gold standard dataset that on average only 10\% wikilinks are suitable for reference inheritance while the rest 90\% are irrelevant. To tackle this imbalance in the dataset, we follow the under-sampling technique based on the repeated edited nearest neighbour method proposed by~\newcite{JMLR:v18:16-365}. 

First, in order to understand the importance of the proposed features we perform a $\chi$-square test for the classification. We report the ranking of the features in Table~\ref{tab:chi}. Subsequently, we plug the features according to this rank from top to bottom one by one into the classifier; the performance obtained thereby are presented in Table~\ref{tab:step1}. We observe that appending the features one by one according to the $\chi$-square test ranking helps to improve the overall performance of the classifier except the last feature which causes no improvements.  

\begin{table}[!tbh]
\begin{center}
    \begin{tabular}{ |p{1.4cm}|p{1cm}|p{1cm}|p{1cm}|p{1cm}|p{1cm}|p{1cm}|p{1cm}|p{1cm}|}
    \hline
    \centering \textbf{Rank} & \centering 1 & \centering 2 & \centering 3 & \centering 4 & \centering 5 & \centering 6 & \centering 7 & \centering 8 \tabularnewline \hline 
    \centering \textbf{Feature} & 
\centering TIS & \centering OSS & \centering ISS & \centering OS & \centering OSVS & \centering VS & \centering ISVS & \centering IS\tabularnewline
\hline
      \end{tabular}
\end{center}
\vspace{-4mm}
\caption{$\chi$-square test ranking of features for Step - I (wikilink classification).}
\label{tab:chi}
\end{table}

\begin{table}[!tbh]
\begin{center}
\begin{small}
    \begin{tabular}{ |p{6.7cm}|p{1.8cm}|p{1.7cm}|p{2cm}|}
    \hline
    \centering \textbf{Features Used} & \centering \textbf{Precision} & \centering \textbf{Recall} & \centering \textbf{F-Measure} \tabularnewline \hline 
    \centering TIS & 
\centering 0.11 & \centering 0.90 & \centering 0.20 \tabularnewline
\hline
    \centering TIS, OSS & \centering 0.16 & \centering 0.36 & \centering 0.20 \tabularnewline
\hline
    \centering TIS, OSS, ISS & \centering 0.39 & \centering 0.37 & \centering 0.33 \tabularnewline
\hline

    \centering TIS, OSS, ISS, OS & \centering 0.47 & \centering 0.44 & \centering 0.40 \tabularnewline
\hline

    \centering TIS, OSS, ISS, OS, OSVS & \centering 0.48 & \centering 0.42 & \centering 0.40 \tabularnewline
\hline

    \centering TIS, OSS, ISS, OS, OSVS, VS &\centering 0.49 & \centering 0.42 & \centering 0.40 \tabularnewline
\hline

    \centering \cellcolor{green} TIS, OSS, ISS, OS, OSVS, VS, ISVS &\centering \cellcolor{green}\textbf{0.50} & \centering \cellcolor{green}\textbf{0.45} & \centering \cellcolor{green}\textbf{0.42} \tabularnewline
\hline

    \centering \cellcolor{green}TIS, OSS, ISS, OS, OSVS, VS, ISVS, IS & \centering \cellcolor{green}\textbf{0.50} & \centering \cellcolor{green}\textbf{0.45} & \centering \cellcolor{green}\textbf{0.42} \tabularnewline
\hline

   \end{tabular}
   \end{small}
\end{center}
\vspace{-3mm}
\caption{Step - I (wikilink classification) evaluation with incremental feature set.}
\label{tab:step1}
\vspace{-3mm}
\end{table}

Some of the representative Wikipedia articles from the CS dataset along with the relevant/irrelevant wikilinks for reference inheritance as correctly predicted by our classifier are noted in Table~\ref{tab:examples_s1}. We observe that the relevant wikilinks are semantically more close to the target Wikipedia articles compared to the irrelevant ones, thus making the former ones as more probable candidates for reference inheritance.

\begin{table}[!tbh]
\vspace{-1mm}
\begin{center}
\begin{small}
    \begin{tabular}{ |p{4cm}|p{6cm}|p{5cm}|}
    \hline
    \centering \textbf{Target Wikipedia article} & \centering \textbf{Relevant  wikilinks} & \centering \textbf{Irrelevant wikilinks} \tabularnewline \hline
    \centering Feature (machine learning)
 & \centering Machine learning, Perceptron & \centering 
Computer vision, Speech recognition\tabularnewline \hline
    \centering Vanishing gradient problem
 & \centering 
Recurrent neural network, 
Back-propagation
 & \centering OPTICS algorithm, Cluster analysis
\tabularnewline \hline
    \centering Sensitivity and specificity
 & \centering 
True positive rate, Precision and recall
 & \centering Clinical research, Airport security
\tabularnewline \hline 
   \end{tabular}
   \end{small}
\end{center}
\vspace{-3mm}
\caption{Representative results of wikilink classification for the CS dataset.}
\label{tab:examples_s1}
\vspace{-3mm}
\end{table}

\subsection{Performance analysis of Step - II}
Table~\ref{tab:step2} reports the performance of our approach for various values of $k$. Considering the difficulty of the task we observe that our system achieves very good performance in terms of precision even for higher values of $k$. Some of the representative recommended references along with the target Wikipedia articles are shown in Table~\ref{tab:examples_s2}. The correct references in the table signify the references which are actually present in the target Wikipedia article's reference section (i.e., in the gold standard dataset). We also observe that the incorrect references which are actually not present in the target Wikipedia article's reference section are not adjudged as relevant by our system. As we cannot compare these references with the gold standard, we perform manual evaluation for only these references in order to analyze how suitable these references are for the target Wikipedia article (see Section~\ref{ME}).

\begin{table}[!tbh]
\vspace{-2mm}
\begin{center}
\begin{small}
    \begin{tabular}{ |p{1.5cm}|p{1.8cm}|p{1.7cm}|p{2cm}|}
    \hline
    \centering $\textbf{k}$ & \centering \textbf{Precision} & \centering \textbf{Recall} & \centering \textbf{F-Measure} \tabularnewline \hline 
    \centering 1 & 
\centering 0.44 & \centering 0.21 & \centering 0.28 \tabularnewline
\hline
    \centering 2 & \centering 0.41 & \centering 0.23 & \centering 0.30 \tabularnewline
\hline
    \centering 3 & \centering 0.40 & \centering 0.26 & \centering 0.30 \tabularnewline
\hline
    \centering 4 & \centering 0.37 & \centering 0.27 & \centering 0.31 \tabularnewline
\hline

    \centering 5 & \centering 0.34 & \centering 0.30 & \centering 0.31 \tabularnewline
\hline
    \centering 10 &\centering 0.25 & \centering 0.35
 & \centering 0.30 \tabularnewline \hline
   \end{tabular}
   \end{small}
\end{center}
\vspace{-3mm}
\caption{Evaluation of Step - II for different values of $k$.}
\label{tab:step2}
\vspace{-2mm}
\end{table}

\begin{table}[!tbh]

\begin{center}
\begin{small}
    \begin{tabular}{ |>{\centering}p{3.5cm}|>{\centering}p{2.5cm}|>{\centering}p{3.4cm}|>{\centering}p{4.5cm}|}
    \hline
    \centering \textbf{Target Wikipedia article} & \centering \textbf{Relevant  wikilink} & \centering \textbf{Correct reference}& \centering \textbf{Incorrect reference} \tabularnewline \hline
    \centering Heap (data structure)
 & \centering Binary heap & \centering 
Introduction to Algorithms & Min-max heaps and generalized priority queues\tabularnewline \hline
    \centering Sensitivity and specificity
 & \centering 
Precision and recall
 & \centering An Introduction to ROC Analysis & Advanced Data Mining Techniques
\tabularnewline \hline
   \end{tabular}
   \end{small}
\end{center}
\vspace{-3mm}
\caption{Representative results of Step - II for the CS dataset.}
\label{tab:examples_s2}
\vspace{-3mm}
\end{table}

\subsection{WikiRef vs the baselines}
\label{bs}
The comparisons of the performances of different baselines with WikiRef are noted in Table~\ref{table:baselines_results}. The table shows that for all the values of $k$ the performance of BL-III and BL-IV is significantly better than BL-I and BL-II respectively leading to the fact that the role of Step - I is very crucial in this context and, in particular, helps boost the performance of the full system. In addition, we observe that the performance we get only using unsupervised ranking in step - II based on any one of the three features (used in BL-III, BL-IV and BL-V respectively) described in section~\ref{Sec-s2} cannot beat the performance of our approach where we use a supervised approach like SVMRank that draws advantage of all the three features.
\begin{table*}[!tbh]
\vspace{-3mm}
\begin{center}
\begin{small}
\begin{tabular} {| >{\centering}p{0.3cm}|>{\centering} p{1.8cm} |>{\centering} p{1cm}|>{\centering}p{1cm} | >{\centering} p{1.3cm}|>{\centering}p{1,3cm} | >{\centering}p{1cm}|>{\centering}p{2.5cm} |}
\hline
$\textbf{k}$ &\textbf{Metric} & \textbf{BL-I}& \textbf{BL-II}& \textbf{BL-III}& \textbf{BL-IV}& \textbf{BL-V}&\textbf{WikiRef}\tabularnewline\hline 
\multirow{3}{*}{1}&Precision&0.22&0.15&\cellcolor{blue}0.38
&0.31&0.36&\cellcolor{green}\textbf{0.44}\tabularnewline
 &Recall&0.04&0.05&\cellcolor{blue}0.18&0.09&0.10&\cellcolor{green}\textbf{0.21}
\tabularnewline
&F-Measure&0.068&0.075&\cellcolor{blue}0.20&0.11&0.14&\cellcolor{green}\textbf{0.28}
\tabularnewline \hline
\multirow{3}{*}{3}&Precision&0.12&0.09&\cellcolor{blue}0.34
&0.26&0.27&\cellcolor{green}\textbf{0.40}
\tabularnewline
 &Recall&0.08&0.1&\cellcolor{blue}0.22&0.17&0.16&\cellcolor{green}\textbf{0.26}
\tabularnewline
&F-Measure&0.096&0.075&\cellcolor{blue}0.28&0.17&0.20&\cellcolor{green}\textbf{0.30}
\tabularnewline \hline
\multirow{3}{*}{5}&Precision&0.1&0.07&\cellcolor{blue}0.31
&0.23&0.25&\cellcolor{green}\textbf{0.34}
\tabularnewline
 &Recall&0.1&0.12&\cellcolor{blue}0.26&0.21&0.22&\cellcolor{green}\textbf{0.30}
\tabularnewline
&F-Measure&0.1&0.08&\cellcolor{blue}0.27&0.18&0.21&\cellcolor{green}\textbf{0.31}
\tabularnewline \hline
\multirow{3}{*}{10}&Precision&0.06&0.06&\cellcolor{green}\textbf{0.25}
&0.22&0.23&\cellcolor{green}\textbf{0.25}
\tabularnewline
 &Recall&0.19&0.19&\cellcolor{blue}0.30&0.25&0.26&\cellcolor{green}\textbf{0.35}
\tabularnewline
&F-Measure&0.09&0.09&\cellcolor{blue}0.26&0.20&0.24&\cellcolor{green}\textbf{0.30}
\tabularnewline \hline
\end{tabular}
\end{small}
\caption{Comparison of the performance of proposed baselines along with WikiRef. Best results are in {green} cells and the most competing baseline results are in {blue} cells.}
\label{table:baselines_results}
\end{center}
\vspace{-3mm}
\end{table*}

\subsection{Manual evaluation}
\label{ME}
The aim of this manual evaluation is to understand the quality of the recommendations returned by WikiRef for the cases which are not present in the gold standard dataset. In the survey, we provide a total of 25 Wikipedia pages from the Computer Science (CS) dataset along with five recommended references which are not already present in the gold standard and ask the annotators to choose the appropriate references which should be there in the reference section of the given Wikipedia article to make it more resourceful. A sample evaluation page can be seen in this link\footnote{\url{https://goo.gl/forms/N0rmN5xPRzhXsoCL2}}.
The list of scientific Wikipedia articles that we use for manual evaluation are noted in Table~\ref{tab:examples_c}. 
10 participants with Computer Science background have taken part in this survey. First, we compute absolute performance of our system which shows on an average, the fraction of references actually found to be relevant by the annotators. We compute the performance score per Wikipage as the fraction of the proposed references found to be relevant by the annotators (averaged over all the annotators). We then compute the average over the full list of 25 Wikipages, obtaining a very good performance score of 0.63 leading to the fact that out of 5 recommended references, close to 3 references are found to be relevant by the annotators. 

Next, we also attempt to correlate the ranking of the competing systems with the implicit ranking imposed by the annotators. For each Wikipedia article, we rank the five references depending on the number of participants voting for a particular reference, using standard fractional ranking method. Then we compute the Spearman's rank correlation coefficient ($\rho$) between these ranks and the ranked list returned by WikiRef as well as other baselines. Table~\ref{table:human_judgements} shows that WikiRef beats the most competitive baseline by a significant margin. 
\begin{table}[!tbh]
\vspace{-2mm}
\begin{center}
    \begin{tabular}{ |p{15cm}|}
    \hline 
    Algebric modeling language, Bayesian model of computational anatomy, Object-based language, Recursive ordinal, Graph isomorphism, Recurrence plot, Ross–Fahroo pseudospectral method, Deterministic finite automaton, Clustal, Graph structure theorem, Bi-directed graph, Algebraic graph theory, Belt machine, Hadwiger–Nelson problem, Finite-state machine, 
Two-phase locking, Urquhart graph, Software transactional memory, Radial basis function kernel, Semi-symmetric graph, Algorithm characterizations, Elliott formula, Just-in-time compilation, Abstract family of languages, PAdES.\\
\hline
       \end{tabular}
\end{center}
\vspace{-4mm}
\caption{Target Wikipedia pages for manual evaluation.}
\label{tab:examples_c}

\end{table}

\begin{table*}[!tbh]
\begin{center}
\begin{tabular} {| >{\centering} p{3cm} |>{\centering} p{1cm}|>{\centering}p{1cm} | >{\centering} p{1.3cm}|>{\centering}p{1.3cm} | >{\centering}p{1cm}|>{\centering}p{2cm} |}
\hline
\textbf{Metric} & \textbf{BL-I}& \textbf{BL-II}& \textbf{BL-III}& \textbf{BL-IV}& \textbf{BL-V}&\textbf{WikiRef}\tabularnewline\hline 
Average $\rho$ &0.099&0.16&0.089&\cellcolor{blue}0.168&
0.104&\cellcolor{green}\textbf{0.203}\tabularnewline\hline 
\end{tabular}
\vspace{-2mm}
\caption{Human judgement experiments comparing the proposed baselines with WikiRef. The best result is highlighted in a {green} cell and the most competing baseline result is highlighted in a {blue} cell.}
\label{table:human_judgements}
\end{center}
\end{table*}

\if{0}
We used one annotator per wikipage, each annotator was an unbiased user, well acquainted with the particular Wikipedia concept page. 
\fi

\subsection{Performance analysis for the Physics (PH) dataset}
So far, we have reported all the performance figures for the Computer science (CS) dataset. In order to investigate the applicability of our system in other domains, we repeat the experiments on the Physics (PH) dataset as well. The results for both step - I and step - II are noted in Table~\ref{table:ph_results}. We use the same eight features (TIS, OSS, ISS, OS, OSVS, VS, ISVS, IS) with Random Forest classifier for step - I and for step - II we again choose the same SVMRank framework fed in with the three features (F1-TI, F1-Vec and F2-TI) as discussed earlier. From Table~\ref{table:baselines_results_pH} we observe that for almost all the cases WikiRef performs better than the baselines.
\begin{table*}[!tbh]
\vspace{-2mm}
\begin{center}
\begin{tabular} {| >{\centering}p{1.2cm}|>{\centering} p{1cm} |>{\centering} p{2cm}|>{\centering}p{2cm} | >{\centering} p{2cm}|}
\hline
 \textbf{Step}& & \textbf{Precision}& \textbf{Recall}& \textbf{F-Measure}\tabularnewline\hline 
Step-I&&0.41&0.44&0.37\tabularnewline\hline
\multirow{6}{*}{Step-II}&k=1&0.45&0.1&0.16
\tabularnewline
 &k=2&0.41&0.13&0.19
\tabularnewline
&k=3&0.38&0.16&0.18
\tabularnewline
 &k=4&0.35&0.20&0.25
\tabularnewline
 &k=5&0.32&0.22&0.23
\tabularnewline
  &k=10&0.30&0.25&0.26
\tabularnewline\hline
\end{tabular}
\vspace{-2mm}
\caption{Performance of WikiRef on the Physics (PH) dataset}
\label{table:ph_results}
\end{center}
\end{table*}

\begin{table*}[!tbh]
\vspace{-2mm}
\begin{center}
\begin{small}
\begin{tabular} {| >{\centering}p{0.3cm}|>{\centering} p{1.8cm} |>{\centering} p{1cm}|>{\centering}p{1cm} | >{\centering} p{1.3cm}|>{\centering}p{1,3cm} | >{\centering}p{1cm}|>{\centering}p{2.5cm} |}
\hline
$\textbf{k}$ &\textbf{Metric} & \textbf{BL-I}& \textbf{BL-II}& \textbf{BL-III}& \textbf{BL-IV}& \textbf{BL-V}&\textbf{WikiRef}\tabularnewline\hline 
\multirow{3}{*}{1}&Precision&0.20&0.24&0.36
&0.36&\cellcolor{blue}0.41&\cellcolor{green}\textbf{0.45}
\tabularnewline
 &Recall&\cellcolor{green}\textbf{0.11}&0.07&0.09&0.09&0.09&\cellcolor{blue}{0.1}
\tabularnewline
&F-Measure&\cellcolor{blue}0.14&0.09&0.12&0.12&0.13&\cellcolor{green}\textbf{0.16}
\tabularnewline \hline
\multirow{3}{*}{3}&Precision&0.18&0.18&0.28
&0.26&\cellcolor{blue}0.29&\cellcolor{green}\textbf{0.38}
\tabularnewline
 &Recall&\cellcolor{green}\textbf{0.18}&0.13&0.15&0.15&0.15&\cellcolor{blue}0.16
\tabularnewline
&F-Measure&\cellcolor{blue}0.16&0.11&0.15&0.15&\cellcolor{blue}0.16&\cellcolor{green}\textbf{0.18}
\tabularnewline \hline
\multirow{3}{*}{5}&Precision&0.15&0.15&0.24
&0.24&\cellcolor{blue}0.25&\cellcolor{green}\textbf{0.32}
\tabularnewline
 &Recall&\cellcolor{green}\textbf{0.22}&0.18&0.19&0.19&0.18&\cellcolor{green}\textbf{0.22}
\tabularnewline
&F-Measure&\cellcolor{blue}0.16&0.12&\cellcolor{blue}0.16&\cellcolor{blue}0.16&\cellcolor{blue}0.16&\cellcolor{green}\textbf{0.23}
\tabularnewline \hline
\multirow{3}{*}{10}&Precision&0.17&0.12&\cellcolor{blue}{0.22}&\cellcolor{blue}0.22&\cellcolor{blue}0.22&\cellcolor{green}\textbf{0.30}
\tabularnewline
 &Recall&\cellcolor{green}\textbf{0.29}&\cellcolor{blue}0.27&0.23&0.23&0.22&{0.25}
\tabularnewline
&F-Measure&0.15&0.13&0.17&\cellcolor{blue}0.18&0.17&\cellcolor{green}\textbf{0.26}
\tabularnewline \hline
\end{tabular}
\end{small}
\vspace{-2mm}
\caption{Comparison of the performance of proposed baselines along with WikiRef for Physics (PH) dataset. Best results are in {green} cells and the most competing baseline results are in {blue} cells.}
\label{table:baselines_results_pH}
\end{center}
\vspace{-5mm}
\end{table*}

\section{Conclusion}
This paper presented a novel two-step recommendation system for enhancing the reference section of the Wikipedia entity pages, dealing with scientific concepts by inheriting the references from the wikilinks present in the page itself. In the first stage we obtain relevant wikilinks for an entity page via a supervised classification approach. In the second stage we recommend a ranked list of references from the relevant wikilinks obtained from the first stage. WikiRef achieves an overall \textit{precision@1} of 0.44 for the CS dataset and 0.45 for the PH dataset. These are very significant improvements over the most competing baselines in both cases. In addition, manual evaluations over 25 wikipages show that WikiRef also recommends the most relevant references that are absent in the gold standard dataset. We have made our code and data publicly available\footnote{\url{https://github.com/KingOfThePirate/Wikiref}}.

Immediate future work would be to focus on further evaluations on various datasets including biology and medicine. Also, the proposed approach can be adapted to filter any type of entity pages' reference section. We further plan to prepare an online tool that triggers WikiRef to recommend references as soon as relevant wikilinks are added to a target Wikipedia article.

\if{0}
\section{Credits}

This document has been adapted from the instructions for  
COLING-2016 proceedings compiled by Hitoshi Isahara and Masao Utiyama,
which are, in turn, based on
the instructions for
COLING-2014 proceedings compiled by Joachim Wagner, Liadh Kelly
and Lorraine Goeuriot,
which are, in turn, based on the instructions for earlier ACL proceedings,
including 
those for ACL-2014 by Alexander Koller and Yusuke Miyao,
those for ACL-2012 by Maggie Li and Michael
White, those for ACL-2010 by Jing-Shing Chang and Philipp Koehn,
those for ACL-2008 by Johanna D. Moore, Simone Teufel, James Allan,
and Sadaoki Furui, those for ACL-2005 by Hwee Tou Ng and Kemal
Oflazer, those for ACL-2002 by Eugene Charniak and Dekang Lin, and
earlier ACL and EACL formats. Those versions were written by several
people, including John Chen, Henry S. Thompson and Donald
Walker. Additional elements were taken from the formatting
instructions of the {\em International Joint Conference on Artificial
  Intelligence}.

\section{Introduction}
\label{intro}

%
%
\blfootnote{
    %
    %
    \hspace{-0.65cm}  
    Place licence statement here for the camera-ready version. See
    Section~\ref{licence} of the instructions for preparing a
    manuscript.
    %
    %
    %
    %
}

The following instructions are directed to authors of papers submitted
to COLING-2018 or accepted for publication in its proceedings. All
authors are required to adhere to these specifications. Authors are
required to provide a Portable Document Format (PDF) version of their
papers. \textbf{The proceedings are designed for printing on A4
  paper.}

Authors from countries in which access to word-processing systems is
limited should contact the publication co-chairs
Xiaodan Zhu
(\texttt{zhu2048@gmail.com}) and Zhiyuan Liu (\texttt{liuzy@tsinghua.edu.cn})
as soon as possible.

We may make additional instructions available at \url{http://coling2018.org/}. Please check
this website regularly.

\section{General Instructions}

Manuscripts must be in single-column format. {\bf Type single-spaced.}  Start all
pages directly under the top margin. See the guidelines later
regarding formatting the first page. The lengths of manuscripts
should not exceed the maximum page limit described in Section~\ref{sec:length}.
Do not number the pages.

\subsection{Electronically-available Resources}

We strongly prefer that you prepare your PDF files using \LaTeX{} with
the official COLING 2018 style file (coling2018.sty) and bibliography style
(acl.bst). These files are available in coling2018.zip 
at \url{http://coling2018.org/}.
You will also find the document
you are currently reading (coling2018.pdf) and its \LaTeX{} source code
(coling2018.tex) in coling2018.zip. 

You can alternatively use Microsoft Word to produce your PDF file. In
this case, we strongly recommend the use of the Word template file
(coling2018.dot) in coling2018.zip. If you have an option, we
recommend that you use the \LaTeX2e{} version. If you will be
  using the Microsoft Word template, you must anonymise
  your source file so that the pdf produced does not retain your
  identity.  This can be done by removing any personal information
from your source document properties.

\subsection{Format of Electronic Manuscript}
\label{sect:pdf}

For the production of the electronic manuscript you must use Adobe's
Portable Document Format (PDF). PDF files are usually produced from
\LaTeX{} using the \textit{pdflatex} command. If your version of
\LaTeX{} produces Postscript files, you can convert these into PDF
using \textit{ps2pdf} or \textit{dvipdf}. On Windows, you can also use
Adobe Distiller to generate PDF.

Please make sure that your PDF file includes all the necessary fonts
(especially tree diagrams, symbols, and fonts for non-Latin characters). 
When you print or create the PDF file, there is usually
an option in your printer setup to include none, all or just
non-standard fonts.  Please make sure that you select the option of
including ALL the fonts. \textbf{Before sending it, test your PDF by
  printing it from a computer different from the one where it was
  created.} Moreover, some word processors may generate very large PDF
files, where each page is rendered as an image. Such images may
reproduce poorly. In this case, try alternative ways to obtain the
PDF. One way on some systems is to install a driver for a postscript
printer, send your document to the printer specifying ``Output to a
file'', then convert the file to PDF.

It is of utmost importance to specify the \textbf{A4 format} (21 cm
x 29.7 cm) when formatting the paper. When working with
{\tt dvips}, for instance, one should specify {\tt -t a4}.

If you cannot meet the above requirements
for the
production of your electronic submission, please contact the
publication co-chairs as soon as possible.

\subsection{Layout}
\label{ssec:layout}

Format manuscripts with a single column to a page, in the manner these
instructions are formatted. The exact dimensions for a page on A4
paper are:

\begin{itemize}
\item Left and right margins: 2.5 cm
\item Top margin: 2.5 cm
\item Bottom margin: 2.5 cm
\item Width: 16.0 cm
\item Height: 24.7 cm
\end{itemize}

\noindent Papers should not be submitted on any other paper size.
If you cannot meet the above requirements for
the production of your electronic submission, please contact the
publication co-chairs above as soon as possible.

\subsection{Fonts}

For reasons of uniformity, Adobe's {\bf Times Roman} font should be
used. In \LaTeX2e{} this is accomplished by putting

\begin{quote}
\begin{verbatim}
\usepackage{times}
\usepackage{latexsym}
\end{verbatim}
\end{quote}
in the preamble. If Times Roman is unavailable, use {\bf Computer
  Modern Roman} (\LaTeX2e{}'s default).  Note that the latter is about
  10\% less dense than Adobe's Times Roman font.

The {\bf Times New Roman} font, which is configured for us in the
Microsoft Word template (coling2018.dot) and which some Linux
distributions offer for installation, can be used as well.

\begin{table}[h]
\begin{center}
\begin{tabular}{|l|rl|}
\hline \bf Type of Text & \bf Font Size & \bf Style \\ \hline
paper title & 15 pt & bold \\
author names & 12 pt & bold \\
author affiliation & 12 pt & \\
the word ``Abstract'' & 12 pt & bold \\
section titles & 12 pt & bold \\
document text & 11 pt  &\\
captions & 11 pt & \\
sub-captions & 9 pt & \\
abstract text & 11 pt & \\
bibliography & 10 pt & \\
footnotes & 9 pt & \\
\hline
\end{tabular}
\end{center}
\caption{\label{font-table} Font guide. }
\end{table}

\subsection{The First Page}
\label{ssec:first}

Centre the title, author's name(s) and affiliation(s) across
the page.
Do not use footnotes for affiliations. Do not include the
paper ID number assigned during the submission process. 
Do not include the authors' names or affiliations in the version submitted for review.

{\bf Title}: Place the title centred at the top of the first page, in
a 15 pt bold font. (For a complete guide to font sizes and styles,
see Table~\ref{font-table}.) Long titles should be typed on two lines
without a blank line intervening. Approximately, put the title at 2.5
cm from the top of the page, followed by a blank line, then the
author's names(s), and the affiliation on the following line. Do not
use only initials for given names (middle initials are allowed). Do
not format surnames in all capitals (e.g., use ``Schlangen'' not
``SCHLANGEN'').  Do not format title and section headings in all
capitals as well except for proper names (such as ``BLEU'') that are
conventionally in all capitals.  The affiliation should contain the
author's complete address, and if possible, an electronic mail
address. Start the body of the first page 7.5 cm from the top of the
page.

The title, author names and addresses should be completely identical
to those entered to the electronical paper submission website in order
to maintain the consistency of author information among all
publications of the conference. If they are different, the publication
co-chairs may resolve the difference without consulting with you; so it
is in your own interest to double-check that the information is
consistent.

{\bf Abstract}: Type the abstract between addresses and main body.
The width of the abstract text should be
smaller than main body by about 0.6 cm on each side.
Centre the word {\bf Abstract} in a 12 pt bold
font above the body of the abstract. The abstract should be a concise
summary of the general thesis and conclusions of the paper. It should
be no longer than 200 words. The abstract text should be in 11 pt font.

{\bf Text}: Begin typing the main body of the text immediately after
the abstract, observing the single-column format as shown in 
the present document. Do not include page numbers.

{\bf Indent} when starting a new paragraph. Use 11 pt for text and 
subsection headings, 12 pt for section headings and 15 pt for
the title. 

{\bf Licence}: Include a licence statement as an unmarked (unnumbered)
footnote on the first page of the final, camera-ready paper.
See Section~\ref{licence} below for details and motivation.

\subsection{Sections}

{\bf Headings}: Type and label section and subsection headings in the
style shown on the present document.  Use numbered sections (Arabic
numerals) in order to facilitate cross references. Number subsections
with the section number and the subsection number separated by a dot,
in Arabic numerals. Do not number subsubsections.

{\bf Citations}: Citations within the text appear in parentheses
as~\cite{Gusfield:97} or, if the author's name appears in the text
itself, as Gusfield~\shortcite{Gusfield:97}.  Append lowercase letters
to the year in cases of ambiguity.  Treat double authors as
in~\cite{Aho:72}, but write as in~\cite{Chandra:81} when more than two
authors are involved. Collapse multiple citations as
in~\cite{Gusfield:97,Aho:72}. Also refrain from using full citations
as sentence constituents. We suggest that instead of
\begin{quote}
  ``\cite{Gusfield:97} showed that ...''
\end{quote}
you use
\begin{quote}
``Gusfield \shortcite{Gusfield:97}   showed that ...''
\end{quote}

If you are using the provided \LaTeX{} and Bib\TeX{} style files, you
can use the command \verb|\newcite| to get ``author (year)'' citations.

As reviewing will be double-blind, the submitted version of the papers
should not include the authors' names and affiliations. Furthermore,
self-references that reveal the author's identity, e.g.,
\begin{quote}
``We previously showed \cite{Gusfield:97} ...''  
\end{quote}
should be avoided. Instead, use citations such as 
\begin{quote}
``Gusfield \shortcite{Gusfield:97}
previously showed ... ''
\end{quote}

\textbf{Please do not use anonymous citations} and do not include
any of the following when submitting your paper for review:
acknowledgements, project names, grant numbers, and names or URLs of
resources or tools that have only been made publicly available in
the last 3 weeks or are about to be made public and would compromise the anonymity of the submission.
Papers that do not
conform to these requirements may be rejected without review.
These details can, however, be included in the camera-ready, final paper.

\textbf{References}: Gather the full set of references together under
the heading {\bf References}; place the section before any Appendices,
unless they contain references. Arrange the references alphabetically
by first author, rather than by order of occurrence in the text.
Provide as complete a citation as possible, using a consistent format,
such as the one for {\em Computational Linguistics\/} or the one in the 
{\em Publication Manual of the American 
Psychological Association\/}~\cite{APA:83}.  Use of full names for
authors rather than initials is preferred.  A list of abbreviations
for common computer science journals can be found in the ACM 
{\em Computing Reviews\/}~\cite{ACM:83}.

The \LaTeX{} and Bib\TeX{} style files provided roughly fit the
American Psychological Association format, allowing regular citations, 
short citations and multiple citations as described above.

{\bf Appendices}: Appendices, if any, directly follow the text and the
references (but see above).  Letter them in sequence and provide an
informative title: {\bf Appendix A. Title of Appendix}.

\subsection{Footnotes}

{\bf Footnotes}: Put footnotes at the bottom of the page and use 9 pt
text. They may be numbered or referred to by asterisks or other
symbols.\footnote{This is how a footnote should appear.} Footnotes
should be separated from the text by a line.\footnote{Note the line
separating the footnotes from the text.}

\subsection{Graphics}

{\bf Illustrations}: Place figures, tables, and photographs in the
paper near where they are first discussed, rather than at the end, if
possible. 
Colour
illustrations are discouraged, unless you have verified that  
they will be understandable when printed in black ink.

{\bf Captions}: Provide a caption for every illustration; number each one
sequentially in the form:  ``Figure 1. Caption of the Figure.'' ``Table 1.
Caption of the Table.''  Type the captions of the figures and 
tables below the body, using 11 pt text.

Narrow graphics together with the single-column format may lead to
large empty spaces,
see for example the wide margins on both sides of Table~\ref{font-table}.
If you have multiple graphics with related content, it may be
preferable to combine them in one graphic.
You can identify the sub-graphics with sub-captions below the
sub-graphics numbered (a), (b), (c) etc.\ and using 9 pt text.
The \LaTeX{} packages wrapfig, subfig, subtable and/or subcaption
may be useful.

\subsection{Licence Statement}
\label{licence}

As in COLING-2014 and COLING-2016, 
we require that authors license their
camera-ready papers under a
Creative Commons Attribution 4.0 International Licence
(CC-BY).
This means that authors (copyright holders) retain copyright but
grant everybody 
the right to adapt and re-distribute their paper 
as long as the authors are credited and modifications listed.
In other words, this license lets researchers use research papers for their research without legal issues.
Please refer to 
\url{http://creativecommons.org/licenses/by/4.0/} for the
licence terms.  

Depending on whether you use British or American English in your
paper, please include one of the following as an unmarked
(unnumbered) footnote on page 1 of your paper.
The \LaTeX{} style file (coling2018.sty) adds a command
\texttt{blfootnote} for this purpose, and usage of the command is
prepared in the \LaTeX{} source code (coling2018.tex) at the start
of Section ``Introduction''.

\begin{itemize}
    %
    %
    \item  This work is licenced under a Creative Commons Attribution 4.0 International Licence. Licence details: \url{http://creativecommons.org/licenses/by/4.0/}
           
    %
    %
    \item This work is licensed under a Creative Commons Attribution 4.0 International License. License details: \url{http://creativecommons.org/licenses/by/4.0/}

\end{itemize}

We strongly prefer that you licence your paper as the CC license
above. However, if it is impossible for you to use that license, please 
contact the publication co-chairs Xiaodan Zhu
(\texttt{zhu2048@gmail.com}) and Zhiyuan Liu (\texttt{liuzy@tsinghua.edu.cn}),
before you submit your final version of accepted papers. 
(Please note that this license statement is only related to the final versions of accepted papers. 
It is not required for papers submitted for review.)

\section{Translation of non-English Terms}

It is also advised to supplement non-English characters and terms
with appropriate transliterations and/or translations
since not all readers understand all such characters and terms.
Inline transliteration or translation can be represented in
the order of: original-form transliteration ``translation''.

\section{Length of Submission}
\label{sec:length}

The maximum submission length is 9 pages (A4), plus an unlimited number of pages for
references. Authors of accepted papers will be given additional space in
the camera-ready version to reflect space needed for changes stemming
from reviewers comments.

Papers that do not
conform to the specified length and formatting requirements may be
rejected without review.

For papers accepted to the main conference, we will invite authors to provide a translation 
of the title and abstract and a 1-2 page synopsis of the paper in a second 
language of the authors' choice. Appropriate languages include but are not 
limited to authors' native languages, languages spoken in the authors' place 
of affiliation, and languages that are the focus of the research presented.

\section*{Acknowledgements}

The acknowledgements should go immediately before the references.  Do
not number the acknowledgements section. Do not include this section
when submitting your paper for review.
\fi

\if{0}

\fi
\end{document}